\documentclass[journal]{IEEEtran}
\IEEEoverridecommandlockouts
\usepackage{cite}
\usepackage{amsmath,amssymb,amsfonts}
\usepackage{algorithmic}
\usepackage{xspace}
\usepackage{graphicx}
\usepackage{array}
\usepackage{textcomp}
\usepackage{xcolor}
\usepackage{tabularx, booktabs}
\usepackage{multirow}
\usepackage{flushend}

\newcommand{\gray}[1]{{\textcolor{gray}{#1}}}

\usepackage{pifont}

\newcommand{\xmark}{\ding{55}}
\newcolumntype{Y}{>{\centering\arraybackslash}X}

\def\BibTeX{{\rm B\kern-.05em{\sc i\kern-.025em b}\kern-.08em
    T\kern-.1667em\lower.7ex\hbox{E}\kern-.125emX}}

\makeatletter
\DeclareRobustCommand\onedot{\futurelet\@let@token\@onedot}
\def\@onedot{\ifx\@let@token.\else.\null\fi\xspace}
\def\eg{\emph{e.g}\@onedot} 

\def\ie{\emph{i.e}\onedot}

\def\etal{\emph{et al}\onedot}
\makeatother

\def\Plus{\texttt{+}}

\usepackage[bookmarks,unicode=true,pdfauthor={Vandad Davoodnia}, pdftitle={SkelFormer: Markerless 3D Pose and Shape Estimation using Skeletal Transformers}]{hyperref}

\hypersetup{
    colorlinks,
    linkcolor={red!85!black},
    citecolor={green!85!black},
    urlcolor={blue!85!black},
}

\usepackage[capitalize]{cleveref}
\crefname{section}{Sec.}{Secs.}
\Crefname{section}{Section}{Sections}
\crefname{table}{Table}{Tables}
\Crefname{table}{Table}{Tables}

\begin{document}

\title{SkelFormer: Markerless 3D Pose and Shape Estimation using Skeletal Transformers}

\author{Vandad~Davoodnia,
        Saeed~Ghorbani,
        Alexandre~Messier,
        and~Ali~Etemad
\thanks{This work was performed during an internship at Ubisoft Laforge partially funded by Mitacs through the Accelerate program.}
\thanks{V. Davoodnia and A. Etemad were with the Department of Electrical and Computer Engineering, Queen's University, Kingston, Ontario, Canada. e-mail: \{vandad.davoodnia,ali.etemad\}@queensu.ca.}
\thanks{V. Davoodnia and S. Ghorbani were with Ubisoft LaForge, Toronto, Ontario, Canada. e-mail: saeed.ghorbani@ubisoft.com}
\thanks{A. Messier was with Ubisoft LaForge, Montreal, Québec, Canada. e-mail: alexandre.messier@ubisoft.com}
}

\maketitle

\begin{abstract}
    We introduce SkelFormer, a novel markerless motion capture pipeline for multi-view human pose and shape estimation. Our method first uses off-the-shelf 2D keypoint estimators, pre-trained on large-scale in-the-wild data, to obtain 3D joint positions. Next, we design a regression-based inverse-kinematic skeletal transformer that maps the joint positions to pose and shape representations from heavily noisy observations. This module integrates prior knowledge about pose space and infers the full pose state at runtime. Separating the 3D keypoint detection and inverse-kinematic problems, along with the expressive representations learned by our skeletal transformer, enhance the generalization of our method to unseen noisy data. We evaluate our method on three public datasets in both in-distribution and out-of-distribution settings using three datasets, and observe strong performance with respect to prior works. Moreover, ablation experiments demonstrate the impact of each of the modules of our architecture. Finally, we study the performance of our method in dealing with noise and heavy occlusions and find considerable robustness with respect to other solutions.
\end{abstract}


\section{Introduction} \label{sec:intro}
Motion capture is an active field of research with applications in sports, entertainment, health, and human-computer interaction. Currently, optical motion capture technology offers the most reliable and accurate solution by using a large number of cameras that detect markers attached to the actor's body. As a result, optical motion capture is costly and has a time-consuming setup process, preventing its effective application in low-budget or outdoor settings. In contrast, markerless optical motion capture offers a more convenient and portable solution for capturing the pose, generally at the cost of accuracy. Therefore, a significant amount of research is dedicated to improving markerless motion capture in recent years, delivering high-quality animations by using only a few RGB cameras \cite{li2021hybrik,yu2022multiview}. Yet, markerless approaches either take a long time to process, \eg, 6 minutes for a 30-second video \cite{rempe2021humor}, or they struggle to perform well in in-the-wild environments \cite{jia2023delving}. Optimization-based solutions that fit a parametric pose or motion model to the detected keypoints often exhibit long run-times, while regression models trained on controlled and in-studio datasets lack generalization due to low-diversity backgrounds, appearance, and lighting conditions \cite{ionescu2013human3}.

In this paper, we propose a novel pipeline for markerless motion capture, which we name SkelFormer. At a high level, SkelFormer consists of two main modules: a 3D keypoint estimator and a skeletal transformer. First, in order to simplify 3D keypoint detection while maintaining generalizability to a wider distribution of scenarios, our method uses a Direct-Linear-Transformation (DLT) \cite{hartley2003multiple} triangulation method on the output of off-the-shelf 2D keypoint estimators trained on in-the-wild data. 

Next, we propose a skeletal transformer motivated by Inverse-Kinematics (IK) approaches to generate body pose and shape parameters rather than relying on the commonly used optimization methods. This module significantly reduces computational overhead while exhibiting more accurate performance. However, the misalignment between the estimated 3D keypoints and the body joint configuration of motion capture data makes the integration of keypoint estimators and our IK module challenging. To address this, we propose a simple joint regressor, trained on a small set of synthetic, and use it to generate synthetic keypoints from motion capture data that are aligned with 2D keypoint estimators. Lastly, we apply several augmentations on the acquired keypoints and train our IK component to predict the human body and shape from the noisy data. To rigorously test the performance of our method, we evaluate SkelFormer in both in-distribution (InD) and out-of-distribution (OoD) settings against prior works while noting that the majority of prior works have been tested in InD settings. Next, detailed ablation experiments demonstrate the impact of each component. Finally, we study the performance of our method and examine its robustness to highly noisy and occluded data. 

In summary, we propose SkelFormer, a regression-based IK solution that converts 3D body keypoint positions to a full-body pose and shape. Our model is able to bridge the gap between the most accurate 3D keypoint detection algorithms and human pose and body mesh estimation with negligible performance degradation. SkelFormer achieves strong results, where we observe that our method outperforms others in InD scenarios. Next, we find strong cross-dataset generalizability through OoD evaluation on two unseen datasets, achieving competitive performance to InD multi-view solutions. Additionally, we show that our method exhibits a higher robustness (less than half of the error of optimization-based solutions), in severe noisy and occluded scenarios.

\section{Related Work}
\subsection{Keypoint Detection}
The 2D keypoint estimation field has seen substantial progress in recent years. Generally, 2D keypoint estimation models are categorized into top-down and bottom-up approaches, each with their trade-offs in speed and accuracy \cite{sun2019deep,cao2017realtime,zhang2020distribution}. Due to the availability of large-scale datasets, such as COCO WholeBody \cite{jin2020whole, lin2014microsoft} and Halpe \cite{fang2022alphapose}, 2D estimators have expanded into whole-body keypoints, potentially impacting 3D human pose and shape estimation. Previous works have proposed several strategies to infer the 3D keypoints of a subject, including semi-supervised learning \cite{mitra2020multiview,davoodnia2022estimating}, temporal \cite{zhu2023motionbert,davoodnia2023human}, and multi-view \cite{iskakov2019learnable,zhang2021adafuse} modeling. PoseBert \cite{zhu2023motionbert} and volumetric Learnable Triangulation (LT) \cite{iskakov2019learnable} are notable examples of temporal and multi-view methods, respectively, reporting 3D keypoint estimation with an error of below pixel-level accuracy. Our method is able to leverage the advances in 3D keypoint estimation by integrating off-the-shelf models into its pipeline.

\subsection{Pose and Shape Estimation}
\textbf{Regression-based} methods generally predict the parameters of a body model, \eg, SMPL \cite{loper2015smpl,romero2022embodied} represented by body shape and pose components. The research on body pose and shape regression can be categorized into single-view and multi-view problems. Single-view approaches generally suffer from the inherent 2D image to 3D pose ambiguities, resulting in worse performance compared to multi-view methods. For instance, Pose2Mesh \cite{choi2020pose2mesh} uses a GraphCNN, consisting of a mesh coarsening encoder-decoder architecture, to regress the human body and shape from a single image. Similarly, GTRS \cite{zheng2022lightweight} proposes a lightweight graph-based transformer network to uplift 2D keypoints to 3D pose and shape parameters. In another work, PyMAF \cite{zhang2021pymaf} also explores feature maps of the visual encoder to perform the regression task. Since our goal is to design a multi-view markerless motion capture pipeline, we only take inspiration from the advances in single-view research \cite{zheng2022lightweight, li2021hybrik}.

In the context of markerless motion capture via multiple views, the majority of methods are supervised, utilizing strategies such as collaborative learning \cite{li20213d}, volumetric feature aggregation \cite{shin2020multi}, multi-view feature fusion via attention \cite{yu2022multiview}, and pixel-aligned feedback fusion \cite{jia2023delving}. Since regression models typically rely heavily on the availability of annotated data and the diversity of postures in the training data, they tend to be limited to in-studio quality and do not perform well on OoD evaluations with background and appearance shifts \cite{jia2023delving}. Although a common solution to this problem is to pre-train the network on in-the-wild datasets, it does not guarantee better generalizability as the neural networks are susceptible to over-fitting and catastrophic forgetting. This is also evident by the best estimation error of 93 \textit{mm} on the in-the-wild Ski-Pose \cite{sporri2016reasearch} dataset reported in a recent work \cite{yu2022multiview}, which is three times bigger than their 33 \textit{mm} error on the Human3.6m \cite{ionescu2013human3} dataset. Our work addresses this limitation in multi-view approaches by incorporating prior knowledge of human pose into the solution using an IK solver trained on a large set of motion capture data, thus improving its generalization to OoD observations.

\textbf{Optimization-based} approaches fit the parameter of the SMPL model to features extracted from an image, such as 2D/3D keypoints and silhouettes \cite{rempe2021humor}. Simplify-x \cite{pavlakos2019expressive} introduced VPoser, a human variational pose prior trained on a large collection of motion capture data, to reduce the complexity of the optimization space. Subsequently, the majority of recent works rely on VPoser to fit the SMPL model to 3D predictions, which is a time-consuming process that can take up to a day to process a one-hour-long video \cite{huang2021dynamic, rempe2021humor}. Furthermore, variational Gaussian models are prone to mean collapse due to the prior distribution assumption. Additionally, as the optimization process is sensitive to initialization, their potential to fit noisy data accurately is hindered \cite{metz2021gradients}. To address this challenge, several works like ProHMR \cite{kolotouros2021probabilistic} have proposed using regression-based models to initialize the optimization variables. Although this approach can speed up and improve accuracy, optimization algorithms remain susceptible to reaching undesirable local minima, especially on noisy, exotic, or unseen poses as discussed in \cite{metz2021gradients}. In summary, though capable of modeling complex movements and interactions, optimization-based models often require careful hyper-parameter tuning for each recording sequence, making them difficult to use in the face of multiple constraints and impractical for fast-paced production. In this paper, we show that compared to optimization solutions, our skeletal transformer is able to obtain a more accurate solutions and exhibits higher robustness to noise and occlusion.

\subsection{Inverse-kinematics Models}
IK is the task of obtaining body joint rotations given several pose constraints, with applications in robotics \cite{csiszar2017solving} and animation \cite{villegas2018neural}. In the context of human pose and shape estimation, HybrIK \cite{li2021hybrik} proposed a hybrid analytical-neural IK solution that obtains the SMPL body rotations given the 3D keypoints estimated from monocular images. They designed their model to disambiguate the 2D image to 3D pose estimation by considering the shape of the human body and breaking the joint rotations to their swing and twist components. However, to our knowledge, IK applications of neural networks have not been explored for multi-view pose and shape estimation. This may be due to the superior performance of optimization methods, such as VPoser \cite{pavlakos2019expressive}, yet at a significant computational cost.

\begin{figure*}[t]
  \centering
  \includegraphics[width=1\textwidth]{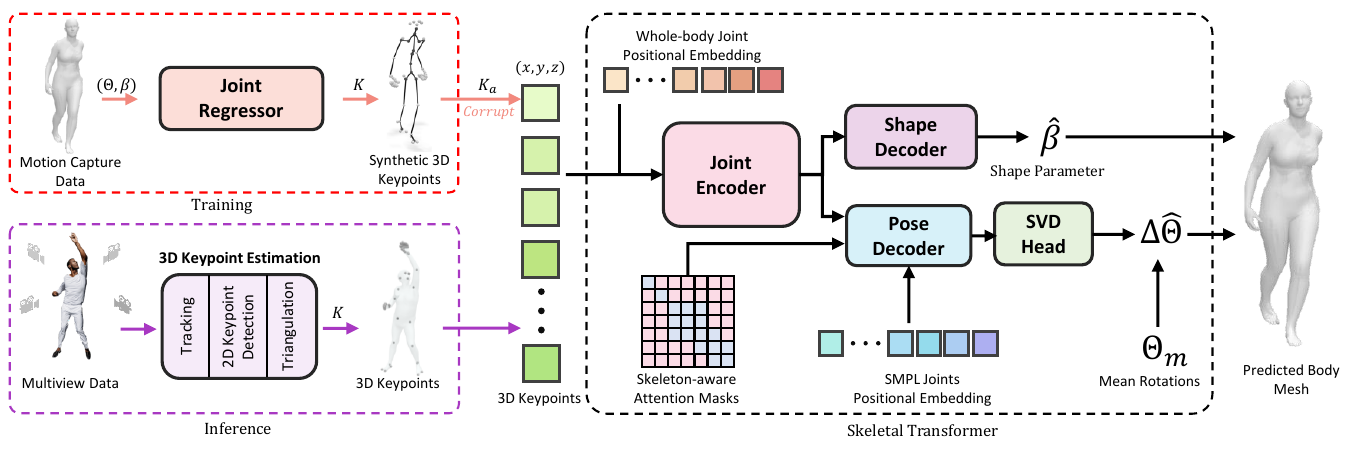}
  \caption{An overview of the proposed skeletal transformer pipeline is demonstrated. During training, noisy 3D keypoints are generated using our joint regressor, while during inference, 3D keypoint are provided by off-the-shelf models. Our proposed skeletal transformer then maps the keypoints onto the SMPL pose and shape parameters.}
  \label{fig:fig2_arch}
\end{figure*}

\subsection{Skeletal Neural Networks}
Previous research has reported superior performance for human motion modeling \cite{raab2023modi}, 3D keypoint refinement \cite{jiang2021skeletor}, and 2D to 3D uplifting \cite{zhu2023motionbert} using skeletal neural networks. In these approaches, the human body structure is modeled as a skeletal graph, and the neural networks exploit the graph structure by learning local and global pose features. Similar techniques have been proposed for hand pose estimation, where the margins for error are much lower than human body pose estimation \cite{zeng2021learning}. Motivated by the recent success of transformers in several fields, such as natural language processing \cite{floridi2020gpt} and computer vision \cite{xu2022vitpose}, they have been used for motion inbetweening and completion \cite{duan2022unified,qin2022motion} of human poses, achieving high-quality results. In this work, we design our model based on a skeletal transformer architecture that can capture the full pose state by providing contextualized latent representations from 3D joint positions.

\section{Methodology}  \label{sec:method}
\noindent \textbf{Overview.} 
As illustrated in \cref{fig:fig2_arch}, our pipeline starts with a 3D keypoint estimator consisting of different sub-modules for tracking, 2D keypoint estimation, and triangulation, for which we use off-the-shelf models. Our proposed skeletal transformer then maps the estimated 3D keypoints onto the SMPL pose and shape parameters. The details of each part are given below.

\subsection{3D Keypoint Estimation} \label{sec:preprocess}

\noindent\textbf{Human Tracking.} We employ Faster R-CNN \cite{ren2015faster}, a well-established object detection model, to track the subjects in the input frames. Although more advanced methods, such as 3D skeleton tracking modules for crowded scenes \cite{bridgeman2019multi}, can be used, we did not observe any misidentification during single-person experiments.

\noindent\textbf{2D Pose Estimation.} To estimate the 2D joints within each frame, we employ HRNet-W48\Plus Dark \cite{sun2019deep, zhang2020distribution} trained on COCO WholeBody dataset \cite{lin2014microsoft}. Since this model is trained on in-the-wild datasets, it helps with the generalizability of our pipeline.

\noindent\textbf{Triangulation.} We choose a simple triangulation method by employing DLT \cite{hartley2003multiple} on 2D keypoints given the extrinsic camera parameters. For this purpose, we consider 2D detection scores for assigning point occlusions.

\begin{figure}[t]
  \centering
  \includegraphics[width=1.0\columnwidth]{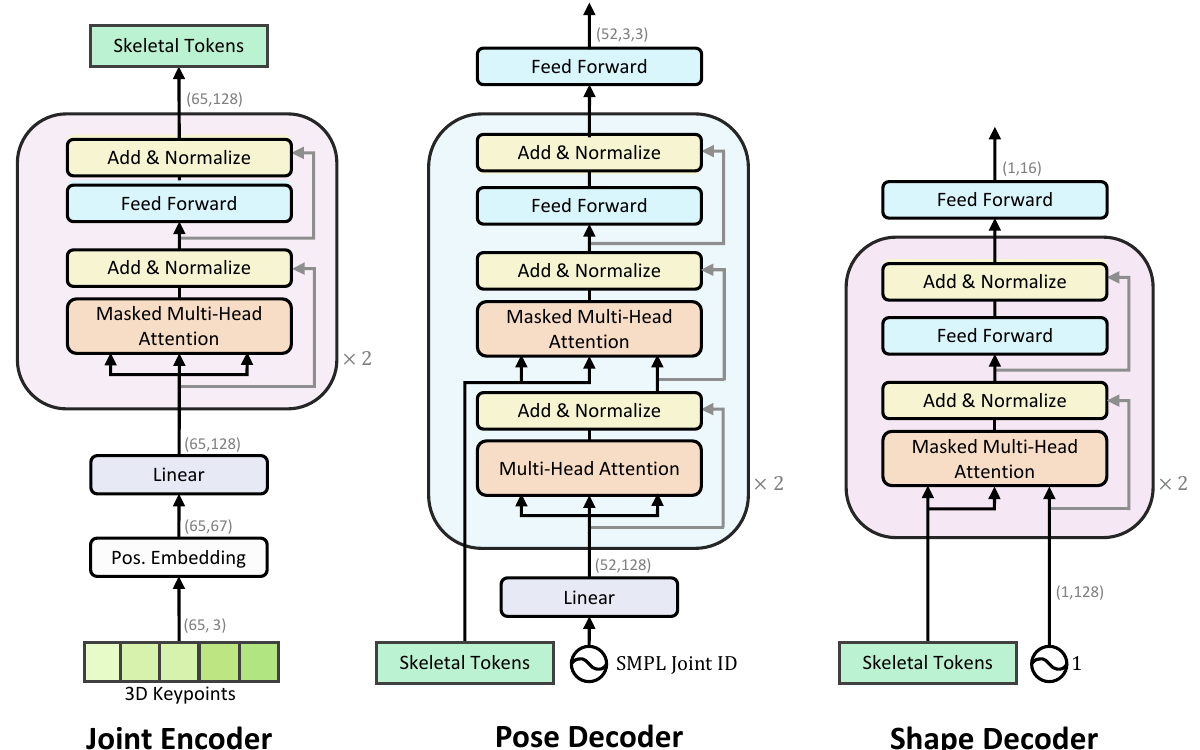}
  \caption{Detailed architectures of our joint encoder, pose decoder, and shape decoder modules are presented.} \label{fig:fig_skeletal_transformer}
\end{figure}

\subsection{Skeletal Transformer}
Traditional IK solvers often assume noise-free constraints, which is not always the case for observations, \eg, 3D keypoints in markerless motion capture. As a result, iterative IK solvers generally perform better than regression models in reaching the local minima given the ground-truth joints, at the cost of additional computations. However, noise and occlusions can cause iterative IK solvers to reach sub-optimal solutions. To address this issue and speed up the process, we introduce an end-to-end learnable pose reconstruction model as illustrated in \cref{fig:fig2_arch}. Following, we present the detailed components of our skeletal transformer.

\noindent\textbf{Joint Encoder.}
As depicted in \cref{fig:fig2_arch}, our network consists of a 3D joint encoder followed by pose and shape decoders. To account for the order of the joints in a skeleton, we first inject information about their ordering by concatenating a positional embedding (\ie, joint ID embedding) vector to each joint input \cite{vaswani2017attention}. The results are then passed through an embedding layer to match the dimensionality of the transformer's hidden layers. Next, we pass the embedded data through a series of transformer encoder blocks consisting of self-attention and feed-forward layers, followed by layer normalization, to obtain the skeletal tokens (see \cref{fig:fig_skeletal_transformer}). To model joint occlusions, we mask out corresponding connections, encouraging the network to use contextual information embedded within the rest of the joints. The result is a latent representation of the joint positions, which is passed to the following decoders. 

\noindent\textbf{Pose Decoder.}
Our pose decoder aims to estimate the body pose given the latent representations of joint positions. A positional embedding is fed into the decoder to relay information about all 52 joints of the SMPL\Plus H skeleton configuration. Then, the skeletal tokens and occlusion masks are passed through the multi-head attention layers in the decoder. Specifically, the decoder consists of several blocks of self-attention, multi-head attention, and feed-forward layers followed by layer normalization (see \cref{fig:fig_skeletal_transformer}). Additionally, to block the unwanted correlations between far-away joints (\eg, left and right hands) that existed in the training set, 
we set the attention weights such that each joint only attends to the other joints that are in a distance of less than 4 nodes in the kinematic tree.

\noindent\textbf{Symmetric Orthogonalization.}
The next step is to obtain the joint rotations from the decoder's output. Although previous works \cite{pavlakos2019expressive,kolotouros2019learning,kolotouros2021probabilistic} have suggested using 6-DoF representations of rotation matrices, we find that SVD-based symmetric orthogonalization proposed in a recent work \cite{levinson2020analysis} yields more accurate results and converges faster. Therefore, the output of the pose decoder is passed through a fully connected residual layer that outputs a square matrix $M_{3 \times 3}$ with SVD of $U \Sigma V^T$. Then, its symmetric orthogonalization $\Theta \in SO(3)$ is obtained by:
\begin{equation}
    \Theta = U \Sigma_o V^T, where \: \Sigma_o = diag(1,1,det(UV^T)).
\end{equation}

\noindent\textbf{Shape Decoder.}
To obtain the shape parameters, we design a shape decoder with a similar architecture to our pose decoder. Since the order of joints does not affect the shape decoder, we remove the self-attention layers and set the sequence length of the shape decoder to one, effectively reducing the decoder to a feed-forward network with multi-head attention (see \cref{fig:fig_skeletal_transformer}).

\subsection{Joint Regressor}
Our goal is to train our skeletal transformer on a large collection of motion capture data consisting of samples represented by SMPL shape and pose parameters. However, the skeleton configuration extracted from whole-body keypoint detectors is not aligned with the SMPL skeletal configuration. To address this, a joint regressor converts SMPL representations to other skeletal configurations. Currently available joint regressors have been reported to be either inaccurate \cite{hedlin2022simple}, or simply obtained by selecting vertices on the \textit{surface} of the body, which is not biomechanically correct nor in accordance with existing body models. Therefore, we propose a novel joint regressor and training scheme to align the SMPL representations to the 3D keypoint configuration using a small amount of synthetic data.

The joint regressor, defined as a linear layer $K=\mathcal{J}V$, is trained to map the body mesh vertices $V\in\mathbb{R}^{6890\times3}$ to 3D keypoints $K\in\mathbb{R}^{J\times3}$, where $J$ is the number of joints in the 3D keypoint configuration. In order to train the joint regressor, we randomly take 10,000 SMPL body samples from the AMASS dataset \cite{mahmood2019amass} and render them from four orthogonal views after removing the root translation and adding random root rotations augmentation. Next, following the process explained in \cref{sec:method}, we use HRNet-W48\Plus Dark \cite{sun2019deep, zhang2020distribution} for whole-body keypoint estimation followed by the DLT triangulation algorithm (\Cref{sec:method}) to obtain the 3D keypoints for each sample. 

To encourage sparsity and avoid out-of-body predictions, previous works \cite{hedlin2022simple} have suggested using an $L_2$ regularization on the joint regressor with the goal of (a) enforcing a sum of 1 for vertex weights of each 3D joint; and (b) encouraging all of the weights of the joint regressor to lie between 0 and 1. However, doing so creates a trade-off between regularization and accuracy. In order to solve this issue, we apply a temperature-scaled Softmax function over the trainable parameters of the joint regressor $\phi$, thus automatically satisfying both constraints. Doing so also gives us control over the vertex sparsity for each of the 3D joints. The joint regressor's weights for the $i_{th}$ keypoint are computed as:
\begin{align}
\mathcal{J}_i(\phi) = \frac{e^{\phi_i / T}}{\sum_{j=1}^{J} e^{\phi_j / T}},
\end{align}
where $T$ controls the sharpness of the distribution of the vertex weights. We use an L-BFGS optimizer \cite{liu1989limited} to increase the training efficiency. We empirically set the temperature to $T=10$ so that 3 to 10 vertices contribute to each joint.

\subsection{Data Preparation}
We extract pairs of 3D keypoints $K$ (using our joint regressor), body joint rotation matrices $\Theta \in \mathbb{R}^{52 \times 3 \times 3}$, and shape parameters $\beta\in\mathbb{R}^{16}$ from human motion capture data. Next, we measure the average joint rotations across the dataset to normalize $\Theta$. As some layers in the skeletal transformer share weights across all the joints, the symmetry of the left and right rotations is important. Yet, the mean pose $\Theta_m$ provided and used in previous works \cite{kanazawa2018end, kolotouros2019learning, kolotouros2021probabilistic} is not symmetrical. To address this, we add the mediolateral mirrored samples to the dataset and use quaternion averaging as described in \cite{markley2007averaging} to measure $\Theta_m$, subsequently normalizing the pose by $\Delta \Theta = \Theta_m^{-1} \Theta$. 

During training, we apply a range of online augmentations to diversify the data and enhance the robustness of our model. These augmentations include:

\noindent\textbf{Masking.} We apply random masking on each joint with a 20\% chance to model partially occluded inputs, simulating scenarios where joints are not visible or considered outliers. For this task, we exploit the inner mechanics of transformers by masking the attention of the occluded joints in the encoder and between the encoder and decoders. 

\noindent\textbf{Rotation.} We randomly rotate the 3D keypoints around the mid-hip point to simulate various orientations of the human body. Specifically, we rotate the points by $\pm180^{\circ}$ along the vertical axis to enhance the model's ability to predict the root rotation accurately. Additionally, we occasionally rotate the body by 90\textdegree~ to simulate lying and sleeping postures, which are often considered a challenging posture in markerless motion capture applications.

\noindent\textbf{Noise Addition.} To model the faulty predictions in the 3D keypoint estimation module, we add random Gaussian noise with a standard deviation proportional to 5\% of the joint annotation confidences provided in the COCO WholeBody dataset \cite{lin2014microsoft}. 

\noindent\textbf{Left-Right Mirroring.} We increase the robustness of our model to mediolateral flips by mirroring the keypoints along the YZ plane by 50\% chance. Since the SMPL model is not symmetrical, we apply the mirroring directly on the inputs and the output of the model before the SMPL FK layer.

\noindent\textbf{Shape Augmentation.} Although AMASS \cite{mahmood2019amass} training set has a large variety of common and exotic poses, it lacks in body shapes variety and includes only 300 different body shapes. In order to mitigate this issue, we randomly augment the body shape parameters before keypoint corruption by an additive Gaussian noise with a standard deviation equal to the standard deviation of all available body shapes.

\noindent\textbf{Outliers.} A common source of noise in 3D triangulation pipelines is heavy 2D keypoint shifts caused by faulty detection in one or more views. Therefore, a straightforward approach is to consider such heavy shifts as outliers and mask those inputs in the network. However, the outlier detection algorithms might fail to detect such instances. In order to increase the robustness of our model to such noises, we apply a large additive Gaussian noise with a standard deviation of one meter to each keypoint with a chance of 1\% during training to increase the robustness of our model to outliers.

\subsection{Training}
In order to train our skeletal transformer, we feed the model with augmented data $K_a$ to obtain pose and shape parameters, which are then passed through the SMPL Forward Kinematics (FK) layer. We then use a combination of rotational, positional, and shape losses to leverage different scopes and granularities to ultimately help with training \cite{sengupta2020synthetic}. Our final loss is calculated as $L=L_{R}+L_{P}+L_{S}$ with details as follows.

\noindent\textbf{Rotation Loss.} We define our rotational loss as the sum of the global and local geodesic distances between the predicted $\hat{\Theta}$ and ground-truth rotations $\Theta$ for each joint: 
\begin{equation}
\label{equ:geo}
L_{R}(\Theta, \hat{\Theta}) = \arccos \left( \frac{\operatorname{tr}(\Theta \hat{\Theta}^\intercal) - 1}{2} \right).
\end{equation}

\noindent\textbf{Position Loss.} 
Our model uses the SMPL FK layer to obtain the keypoints $\hat{K}$ and vertices $\hat{V}$. We then define a positional loss using $L_{P}=L_{1;s}(\hat{K}, K) + L_{1;s}(\hat{V}, V)$, where the $L_{1;s}$ represents a smoothed $L_1$ loss \cite{girshick2015fast}. This loss can be seen as a combination of $L_1$ and $L_2$ distances, which is less susceptible to outliers compared to $L_2$ with less near-zero penalty compared to the $L_1$ distance.

\noindent\textbf{Shape Loss.} 
Finally, we minimize the $L_{S}=L_2(\hat{\beta}, \beta)$ distance between the estimated $\hat{\beta}$ and ground-truth shape parameters $\beta$ to train our shape decoder.

\section{Experiments}

\subsection{Datasets} 
\noindent\textbf{AMASS.} 
The Archive of Motion Capture as Surface Shapes (AMASS) \cite{mahmood2019amass} is a collection of 3D human pose and shape information collected from multiple motion capture databases. It contains over 40 hours of motion capture data from more than 300 subjects and spans over 11,000 actions. We follow the standard train, test, and validation splits used in prior works \cite{rempe2021humor, pavlakos2019expressive}. This dataset is used solely for training the skeletal transformer IK solver.  

\noindent\textbf{Human3.6m.}
Human3.6m \cite{ionescu2013human3} is the standard benchmark for evaluating 3D human pose, shape, and body estimation in multi-view and single-view approaches. Following previous works \cite{moon2022neuralannot}, we perform our evaluations using Protocol-I, where the root-centered MPJPE and PA-MPJPE on subjects 9 and 11 are measured. We use this dataset for \textit{\textbf{InD}} evaluation of our method.

\noindent\textbf{RICH.}
Real scenes, Interaction, Contact and Humans (RICH) dataset \cite{huang2022capturing} is a recently published dataset of multi-view videos with accurate markerless motion-captured bodies and scenes. The test set contains one withheld scene and 7 unseen subjects in 52 scenarios, captured using four cameras. We use this dataset for \textit{\textbf{OoD}} evaluation of our method in outdoor settings.

\noindent\textbf{MPI-INF-3DHP.}
The Max Planck Institute for Informatics 3D Human Pose dataset (MPI-INF-3DHP) \cite{mehta2017monocular} is a collection of over 1.5 million frames captured from eight angles, featuring eight actors performing various actions like sitting, jumping, dancing, and exercising. We follow the previous works by evaluating our model on subject 8 of the training set \cite{shin2020multi, liang2019shape}. We use this dataset for \textit{\textbf{OoD}} evaluation of our method.

\subsection{Evaluation Metrics}
In order to evaluate the performance of our method, we employ standard evaluation metrics in 3D pose estimation literature. Mean-Per-Joint-Position-Error (\textbf{MPJPE}) measures the Euclidean distance between the estimated joint positions and the ground-truth joint positions, averaged over all joints in the skeleton. 
\textbf{PA-MPJPE} is an extension of MPJPE, where a rigid alignment between the estimated and ground-truth keypoints is applied prior to error measurement. This metric shows how well the skeleton is estimated, regardless of scaling and rotation. Additionally, we report \textbf{MPVPE} and \textbf{PA-MPVPE} to show the error between ground-truth and predicted vertices of body mesh. Additionally, we report \textbf{AUC} and \textbf{PCK} at a threshold of 150 \textit{mm} according to MPI-INF-3DHP \cite{mehta2017monocular} evaluation criteria. Finally, \textbf{Rotation Error} is measured by the geodesic distance between the ground-truth and predicted poses.

\subsection{Implementation Details} \label{appx:imp_detail}
\noindent\textbf{Hyper-parameters.}
We select 65 joints from the 133 joints of the whole-body skeleton configuration, excluding most facial landmarks while keeping the eyes, ears, and nose. We choose two transformer blocks for the encoder and each of the decoders. As illustrated in \cref{fig:fig_skeletal_transformer}, We use a positional embedding of size 64 in the encoder and decoder while setting hidden layer dimensions to 128 within all layers. The shape and pose decoder heads inherit 1024 dimensional residual layers.

\noindent\textbf{Optimization.}
We train our model using AdamW optimizer \cite{loshchilov2017decoupled} with a batch size of 1024 on an NVIDIA A4000 GPU. We chose a learning rate of 1e-3, which is warmed up with a factor of 1e-4 for the first 2000 iterations and gradually reduced with a cosine annealing scheduler over 50000 iterations until it reaches 1e-7. The training of the network takes less than 18 hours to complete.

\noindent\textbf{Mirror Test.}
Similar to the flipping test performed in 2D keypoint estimation methods \cite{xu2022vitpose,sun2019deep}, we perform a mirroring test during inference to reduce the model's biases towards left and right body parts.

\noindent\textbf{Computation Cost.}
Our model contains 6.631~\textit{M} parameters with 159.482~\textit{M} FLOPs for a single input. As a result, our skeletal transformer can solve the IK problem in 66~\textit{ms} for a batch size of 512 using approximately 4G of GPU memory. Consequently, our SkelFormer pipeline takes 274~\textit{ms} to predict pose and shape parameters from one frame of four cameras.

\begin{table}[t]
  \centering
  \caption{The comparison of our method in InD settings against prior multi-view works on the full test set of the Human3.6m dataset. * denotes the results from using ground-truth 3D keypoints as input}
  \label{tab:tab3_h36m}
  \setlength
  \tabcolsep{3pt}
  \scriptsize
  \begin{tabular}{lccc}
    \hline    
    \textbf{Method} & \textbf{MPJPE}$\downarrow$ & \textbf{PA-MPJPE}$\downarrow$ & \textbf{Output}  \\
    \hline
    \hline
    \gray{CPN\Plus DLT} \cite{chen2018cascaded}            & \gray{32.1}  & \gray{27.8}   & \gray{Joint Pos. Only}   \\
    \gray{LT}  \cite{iskakov2019learnable}            & \gray{20.7}  & \gray{17.0}   & \gray{Joint Pos. Only}     \\
    \gray{Pose2Mesh} \cite{choi2020pose2mesh}*        & \gray{29.0} & \gray{23.0}    & \gray{Mesh Only} \\
    Huang \etal \cite{huang2017towards}              & 58.2  & 47.1  & Joint Rot.\Plus Mesh     \\
    Shin and Halilaj (SPIN$^{{4,cal}}$) \cite{shin2020multi}   & 49.8  & 35.4  & Joint Rot.\Plus Mesh   \\
    Shin and Halilaj (main) \cite{shin2020multi}      & 46.9  & 32.5  & Joint Rot.\Plus Mesh   \\
    Gong \etal \cite{gong2022progressive}            & 53.8  & 42.4  & Joint Rot.\Plus Mesh  \\
    Jiang \etal \cite{jiang2022multi}                & 50.2  & 37.3  & Joint Rot.\Plus Mesh \\     
    Jia \etal \cite{jia2023delving}                  & 33.0  & 26.9  & Joint Rot.\Plus Mesh  \\
    SMPLify-X (LT) \cite{pavlakos2019expressive}      & 26.3  & 21.2   & Joint Rot.\Plus Mesh  \\
    SkelFormer (CPN)                                  & 33.5 & 27.8     & Joint Rot.\Plus Mesh  \\
    SkelFormer (LT)                          & \textbf{25.2}   & \textbf{20.6}   & Joint Rot.\Plus Mesh  \\
    \hline
  \end{tabular}
\end{table}

\begin{table}[t]
\centering
  \caption{Comparison of our method in OoD setting against prior works. }
  \label{tab:tab5_mpi_neu_ann}
  \setlength
  \tabcolsep{3pt}
  \scriptsize
  \begin{tabular}{lcccccc}
    \hline    
    \textbf{Method} & \textbf{MPVPE}$\downarrow$  & \textbf{MPJPE}$\downarrow$  &\textbf{PA-MPJPE}$\downarrow$  &  \textbf{PCK}$\uparrow$ & \textbf{AUC}$\uparrow$ & \textbf{OoD} \\
    \hline
    \hline
    \multicolumn{7}{c}{RICH Dataset} \\
    \hline
    METRO \cite{lin2021end}               &    134.5 & 129.6 & -   & - & -   &    \checkmark     \\
    METRO \cite{lin2021end}               &    107.9 & 98.8 & -  & - & -   &    \xmark     \\
    SA-HMR \cite{shen2023learning}        &    103.0 & 93.9 & -    & - & -  &    \xmark      \\
    IPMAN-R \cite{tripathi20233d}         &   89.9  &       79.0     &    47.6   & - & -    &   \xmark        \\
    SPIN \cite{kolotouros2019learning}    &    129.5  &      112.2     &   71.5  & - & -     &    \checkmark       \\
    PARE \cite{kocabas2021pare}           &   125.0 &      107.0     &    73.1    & - & -    &     \checkmark  \\
    CLIFF \cite{li2022cliff}              &    122.3 &       107.0    &    67.2   & - & -     &    \checkmark             \\
    SkelFormer (HRNet)                    & \textbf{39.9} & \textbf{44.2}  & \textbf{35.6} & - & -    &      \checkmark  \\ 
    \hline
    \multicolumn{7}{c}{MPI-INF-3DHP Dataset} \\
    \hline
    Liang and Lin \cite{liang2019shape}       &  - & -          & 59.0    & 95.0   & 65.0      &   \xmark  \\
    Shin and Halilaj \cite{shin2020multi}   &    - & -           & 50.2   & 97.4    & 65.5      &   \xmark  \\
    Jia \etal \cite{jia2023delving}    & - & -   & \textbf{48.4}   & \textbf{98.6}    &   67.3   &   \xmark   \\
    SkelFormer (HRNet)                   &   - & -        & 54.8 & 97.5 & \textbf{67.4}        &    \checkmark \\
    \hline
  \end{tabular}
\end{table}

\subsection{Results}
\subsubsection{Performance}
\noindent\textbf{InD Testing.}
To evaluate SkelFormer, we first compare its performance to prior works on the Human3.6m \cite{ionescu2013human3} dataset. In this experiment, most benchmarks pre-train their models on multiple datasets and fine-tune them on the Human3.6m training set. Accordingly, we report the performance of our model using keypoint estimators, \ie, CPN \cite{chen2018cascaded} and LT \cite{iskakov2019learnable}, trained on Human3.6m dataset (InD). To this end, we use 3D predictions from CPN (followed by DLT triangulation \cite{hartley2003multiple}) and LT, both of which are trained following the evaluation protocol described in \cite{ionescu2013human3}. Moreover, we train our skeletal transformer on the AMASS dataset \cite{mahmood2019amass} using a 17-joint configuration as per Human3.6M. Delving deep into the results in \cref{tab:tab3_h36m}, we observe that our method outperforms the best regression model \cite{jia2023delving}. Additionally, SkelFormer significantly outperforms optimization solutions from 2D multi-view keypoints and from 3D keypoint reported in \cite{huang2017towards} and SMPLify-X (LT) \cite{pavlakos2019expressive}. A notable solution is Pose2Mesh \cite{choi2020pose2mesh}, another regression-based approach that exploits the 3D ground-truth joints as input. However, our method can still outperform Pose2Mesh despite not using the ground-truth information. Finally, although not directly comparable, our method can achieve the closest performance to solutions that only predict joint positions, namely, LT \cite{iskakov2019learnable} and CPN\Plus DLT~\cite{chen2018cascaded}.

\begin{figure*}[t]
  \centering
  \includegraphics[width=1\textwidth]{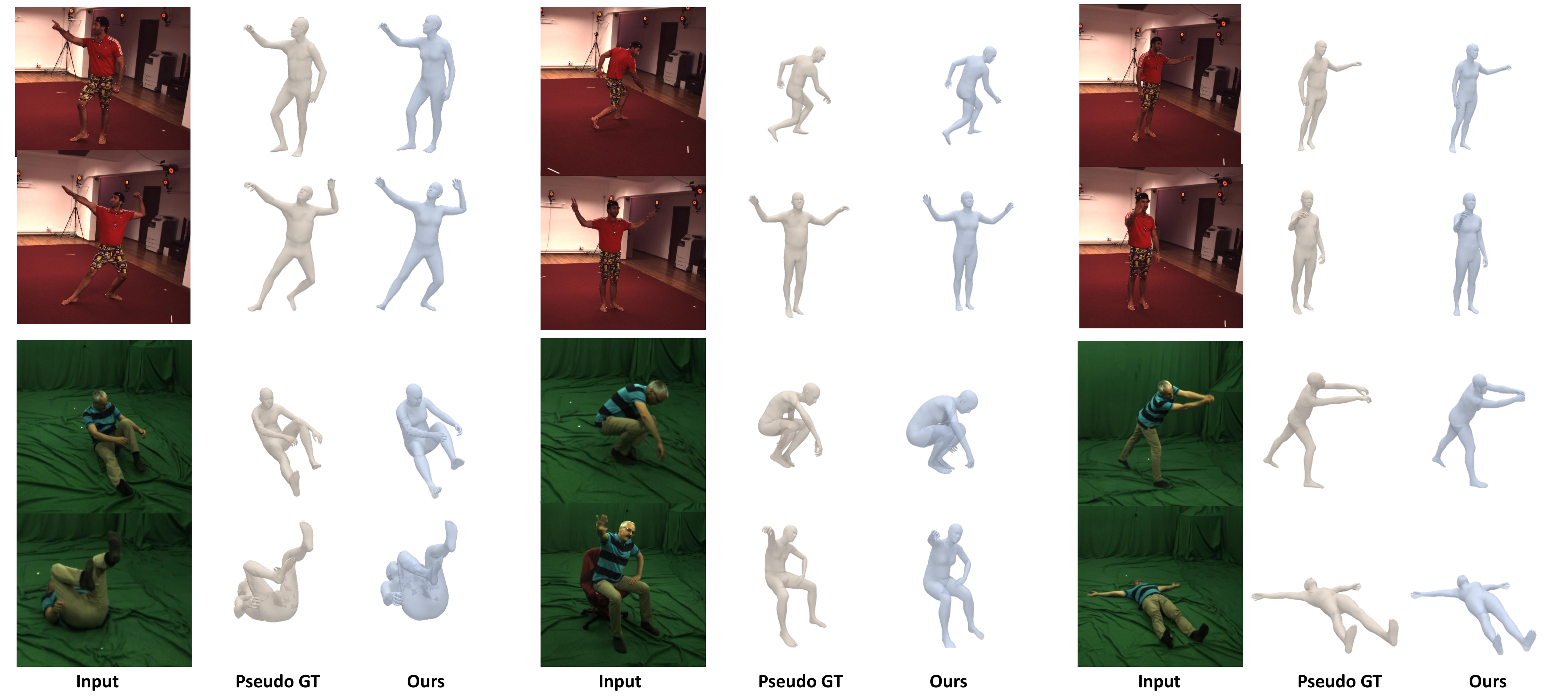}
  \caption{A visual comparison with the pseudo-ground-truth from \cite{moon2022neuralannot} is provided, presenting the realism and accuracy of our SkelFormer.} 
  \label{fig:fig5_hm_mpi_samples}
\end{figure*}

\noindent\textbf{OoD Testing.}
Next, to test our method in OoD settings, we use HRNet-W48\Plus Dark \cite{sun2019deep, zhang2020distribution} as the keypoint extractor, which has not been trained on the RICH \cite{huang2022capturing} or the MPI-INF-3DHP \cite{mehta2017monocular} datasets. Moreover, we keep our entire pipeline frozen and do not fine-tune any of its components on any portion of these datasets. The results are presented in \cref{tab:tab5_mpi_neu_ann}. It should be noted that while prior works on the RICH dataset are monocular pose estimation approaches, all prior works on MPI-INF-3DHP are multi-view solutions and use all available views. On the RICH dataset, \cref{tab:tab5_mpi_neu_ann} shows that other solutions, such as SPIN \cite{kolotouros2019learning} and CLIFF \cite{li2022cliff}, suffer greatly in OoD setups (obtaining 172\% and 127\% additional error w.r.t. their InD performance on Human3.6m dataset). In contrast, our method shows overall competitive results compared to InD solutions by outperforming prior works on RICH while showing relatively minor degradation on the MPI-INF-3DHP dataset. Finally, we demonstrate the fitting quality of our model compared to the pseudo-ground-truth provided in \cite{moon2022neuralannot} on the Human3.6m and the MPI-INF-3DHP datasets in \cref{fig:fig5_hm_mpi_samples}, showing improvements in the feet and hands regions overall. 


\begin{table}[t]
  \caption{Ablation Study results in the presence of 20\% occlusion and a Gaussian noise of $\sigma = 20$ mm.}
  \label{tab:tab6_ablation}
  \setlength
  \tabcolsep{3pt}
  \scriptsize
  \begin{tabular}{lccc}
    \hline
    \textbf{Experiments} &    \textbf{MPVPE}$\downarrow$ &\textbf{PA-MPVPE}$\downarrow$ & \textbf{Rot. Error}$\downarrow$  \\
    \hline
    \hline    
    \textbf{SVD Symmetric Orthogonalization}         & \textbf{18.6}  & \textbf{12.1}    & \textbf{2.94}     \\ 
    6-DoF Symmetric Orthogonalization                & 19.2  & 12.6    & 3.13     \\ 
    \hline
    Local Rot. Loss                   & 19.9  & 13.3    & 3.01     \\ 
    Global Rot. Loss                   & 19.3  & 13.2   & 3.63     \\ 
    \textbf{Global\Plus Local Rot. Loss}  & \textbf{18.6}  & \textbf{12.1}    & \textbf{2.94}     \\ 
    \hline
    Part-based Decoder Att. Weights      & 18.8  & 12.1   & 2.96  \\ 
    $d=1$ Decoder Att. Weights           & 19.6  & 12.8   & 2.97  \\ 
    $d=2$ Decoder Att. Weights           & 19.1  & 12.4   & 2.93  \\ 
    $d=3$ Decoder Att. Weights           & 19.3  & 12.6   & 2.96  \\ 
    \textbf{$d=4$ Decoder Att. Weights}  & \textbf{18.6}  & \textbf{12.1}   & \textbf{2.94}  \\ 
    $d=5$ Decoder Att. Weights           & 18.9  & 12.6   & 2.97  \\ 
    w/o Decoder Att. Weights             & 19.7  & 12.8   & 3.04  \\ 
    \hline
    w/o Rotation Normalization           & 19.5  & 13.1   & 3.01  \\ 
    w/o Mirror Test                      & 27.7  & 19.7   & 3.56  \\ 
    w/o Shape Aug.                       & 19.8  & 12.8   & 2.92     \\ 
    \hline
  \end{tabular}
\end{table}

\subsubsection{Ablation Study}
We test the importance of different network components and report the results in \cref{tab:tab6_ablation}. Given our goal of increasing generalizability in the presence of noise and occlusions, we conduct experiments on motion capture data from the AMASS \cite{mahmood2019amass} testing set at the presence of 20\% occlusion and additive Gaussian noise with $\sigma=20$ \textit{mm}. First, we demonstrate the effectiveness of symmetric orthogonalization by replacing our SVD operation with the commonly used 6-DoF representation, showing a drop of 0.8 \textit{mm} and 0.19\textdegree ~of MPVPE and rotational error when SVD is removed. Next, we experiment with different combinations of local and global (after FK) rotational loss functions to train our model. We observe that combining global and local rotation losses results in a better performance. Next, we showcase the effectiveness of our decoder masking strategy using the attention weights in three experiments: \textit{i}) part-based experiment, where we restrict the attention within upper right, upper left, lower right, lower left, and center regions of the body; \textit{ii}) node distance experiments, where we restrict the attention based on the node distance $d$ in the skeleton kinematic tree with values between $1$ and $5$; and \textit{iii}) without skeleton-aware attention masks, where we allow each joint to attend to any other joints in the decoder. We observe that the best results are obtained for $d=4$, improving over the vanilla transformer decoder by 1.1 \textit{mm} MPVPE and 0.2\textdegree ~rotation error. Lastly, we show the impact of rotation normalization, mirror testing, and shape augmentation, where a significant drop in performance is seen in the individual absence of each of these components, highlighting their significance.

\begin{figure}[t]
  \centering
  \includegraphics[width=\columnwidth]{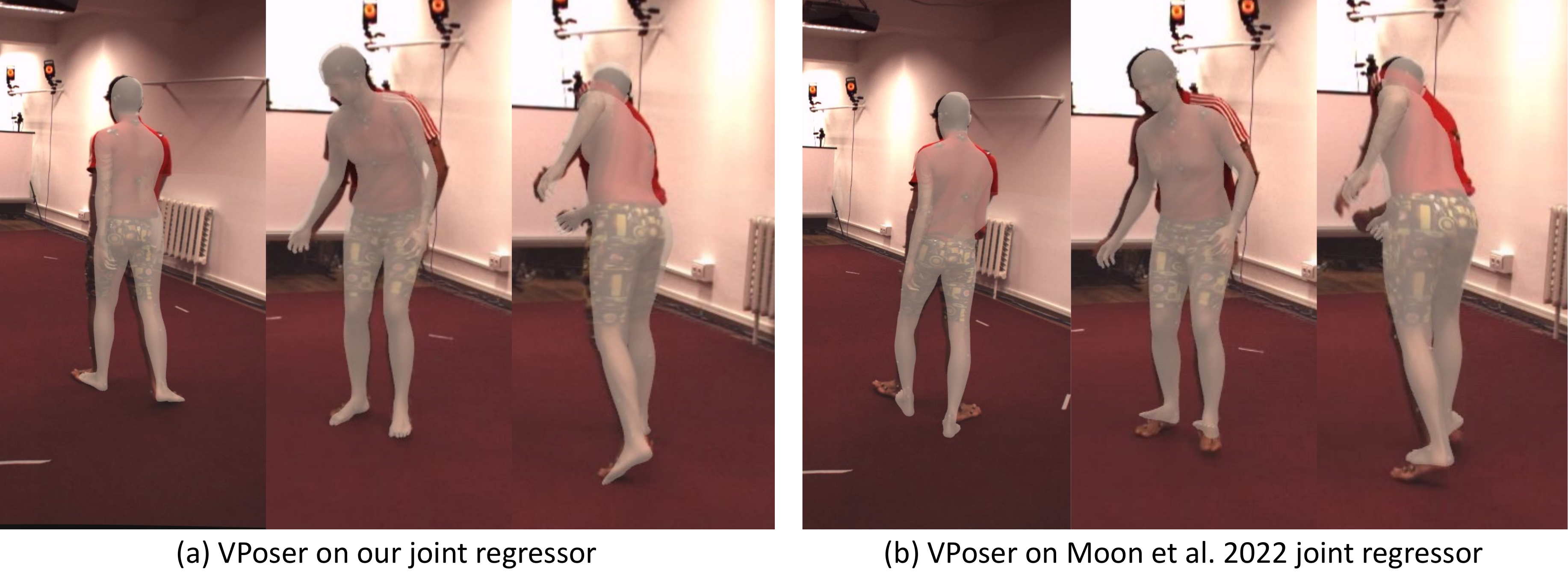}
  \caption{The fitting performance of VPoser is demonstrated while using (a) our proposed joint regressor; and (b) the joint regressor from Moon \etal \cite{moon2022neuralannot}.}
  \label{fig:fig3_regressor}
\end{figure}

Finally, we perform a qualitative experiment on our joint regressor. To this end, we use VPoser \cite{pavlakos2019expressive} to fit onto 3D keypoints using our joint regressor and the one provided in prior works \cite{pavlakos2019expressive} on Human3.6m dataset \cite{ionescu2013human3}. In \cref{fig:fig3_regressor}, we show the visual fidelity of the effect of our joint regressor in comparison to \cite{moon2022neuralannot}, specifically, in better fitting to the feet regions.

\begin{figure}[t]
  \centering
  \includegraphics[width=\columnwidth]{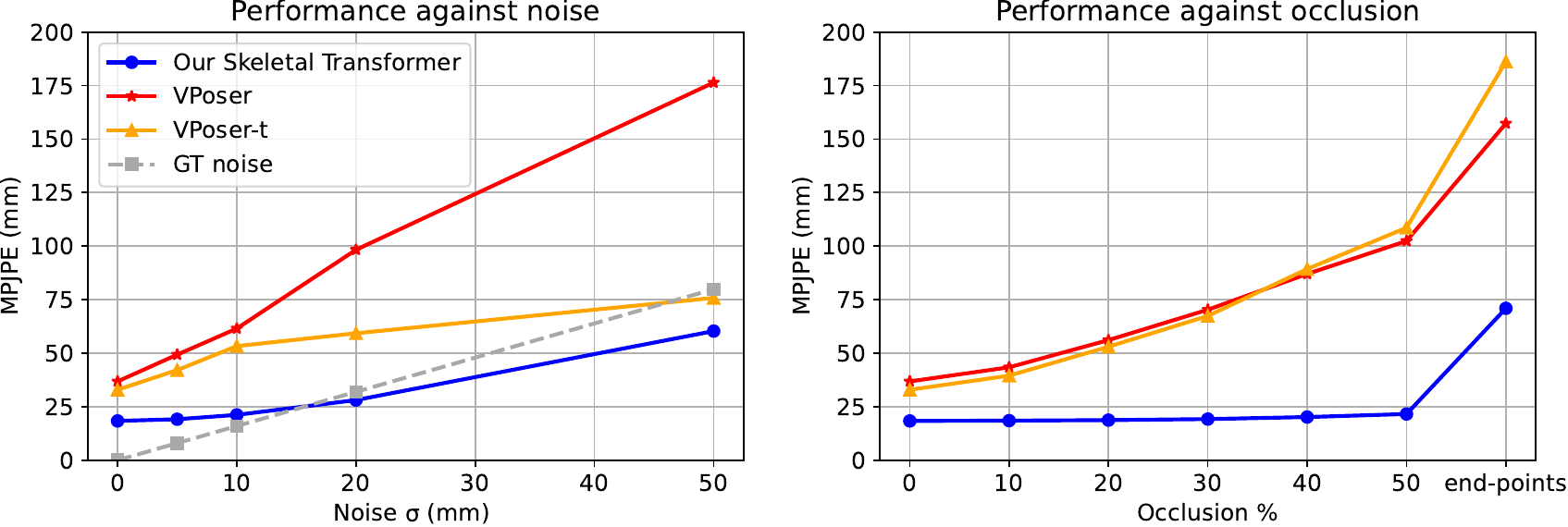}
  \caption{Robustness of our skeletal transformer is highlighted in the presence of different levels of noise and occlusion by comparing it against VPoser and VPoser-t.}
  \label{fig:fig35_noise_occlusion}
\end{figure}

\subsubsection{Robustness to Noise and Occlusions}
In order to evaluate the performance of our model on motion capture data, we perform experiments by simulating noise and occlusion. For these experiments, we compare our skeletal transformer with VPoser \cite{pavlakos2019expressive} and its temporal version, called VPoser-t \cite{rempe2021humor}, which tries to maximize temporal consistency during optimization. \Cref{fig:fig35_noise_occlusion} demonstrates the performance of different models on the AMASS \cite{mahmood2019amass} testing set. In the first experiment, we introduce varying noise levels to the input data and evaluate the robustness of our method. More specifically, Gaussian noise with varying standard deviations up to 50 \textit{mm} is added to the input, effectively increasing the MPJPE of ground truth up to 80 \textit{mm} (referred to as GT noise). However, our skeletal transformer predicts body pose and shape parameters, which result in lower error after $\sigma=15$ \textit{mm}. Additionally, its performance only degrades by 19.7 \textit{mm} at maximum noise level, while VPoser fails to predict less noisy poses. Lastly, by comparing our method to a temporal model (VPoser-t), we demonstrate the model's robustness to noisy scenarios.

\begin{figure*}[t]
  \centering
  \includegraphics[width=0.9\linewidth]{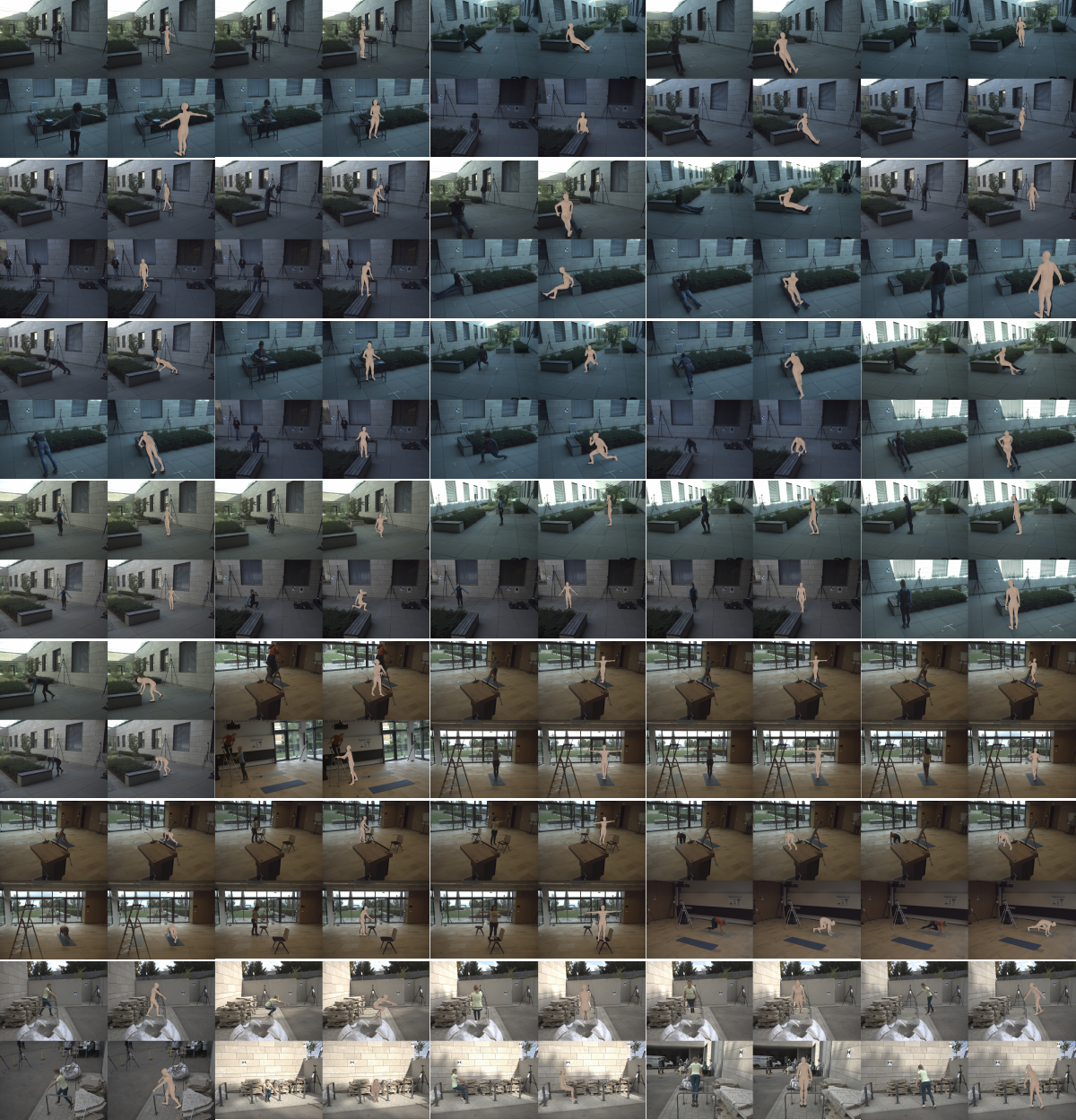}
  \caption{Sample results on RICH dataset are presented.} 
  \label{fig:fig7_rich}
\end{figure*}

\begin{figure*}[!ht]
  \centering
  \includegraphics[width=0.95\linewidth]{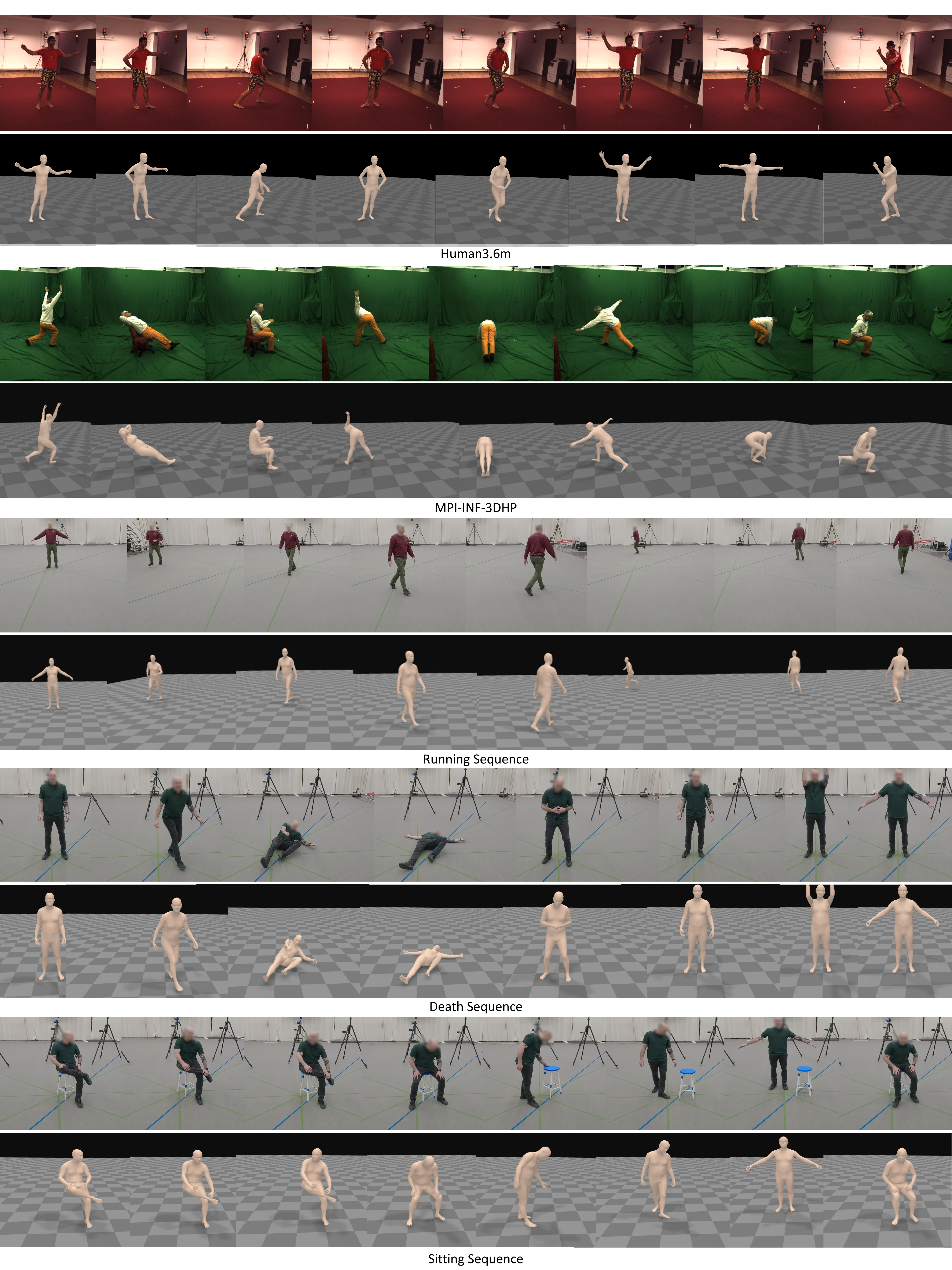}
  \caption{Sample results on Human3.6m, MPI-INF-3DHP, and our collected videos are presented.} 
  \label{fig:fig6_qualitative}
\end{figure*}

\begin{figure*}[!ht]
  \centering
  \includegraphics[width=0.9\textwidth]{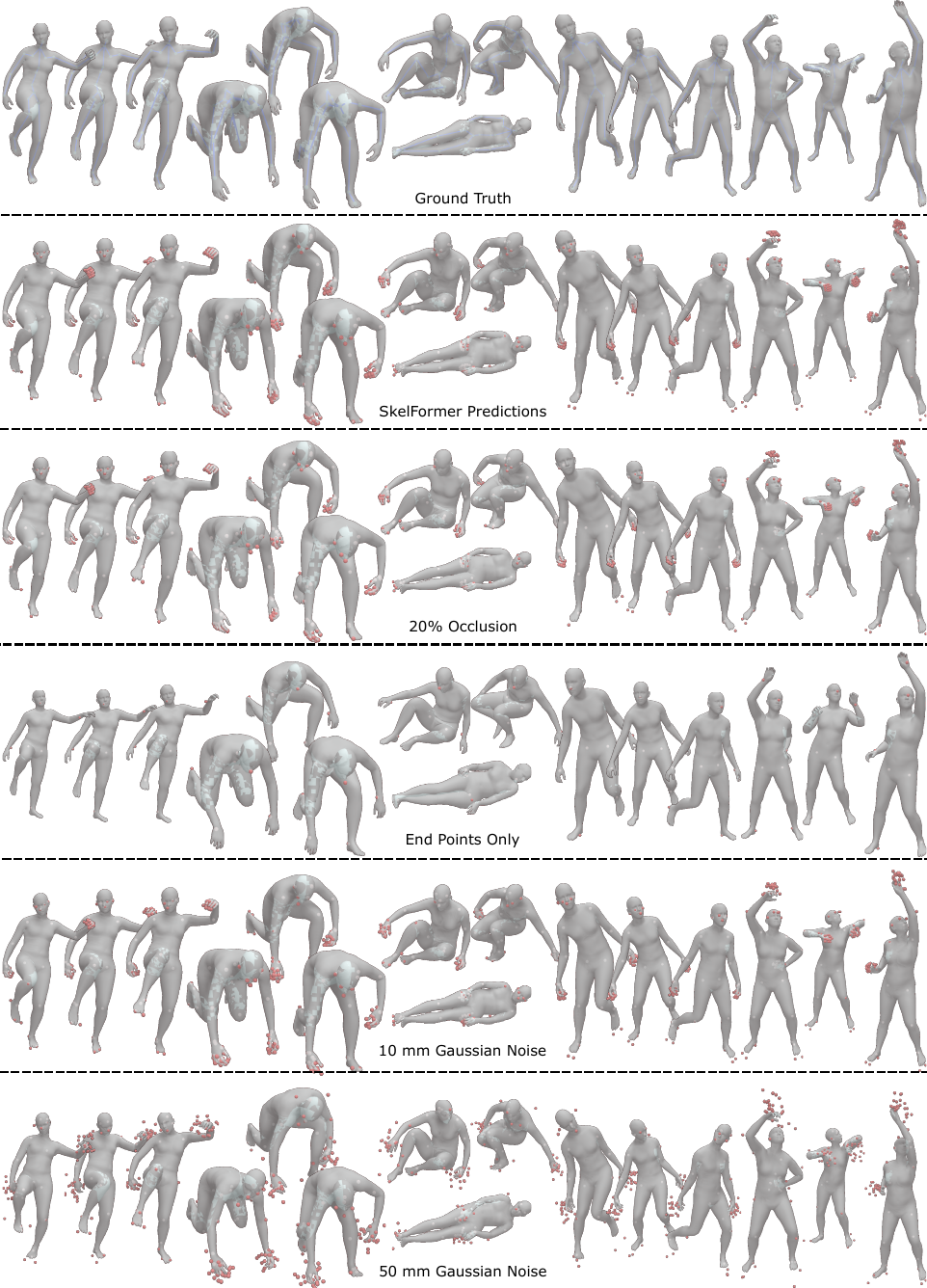}
  \caption{We demonstrate the predictions of SkelFormer on its worst-performing sequences during the occlusion and noise experiments. The red points represent the model's inputs.} 
  \label{fig:fig4_amass}
\end{figure*}

Next, we report the performance of our model in the presence of occlusions, where each joint is randomly masked with varying occlusion amounts of up to 50\%. \Cref{fig:fig35_noise_occlusion} shows that the performance of our skeletal transformer barely changes between 0 and 50\% occlusion, thus showing the model's capability to exploit local and global joint information. In contrast, VPoser's performance drops by more than 20 \textit{mm} in the presence of only 20\% occlusion. We finally report the performance in an extreme occlusion scenario, often seen in IK applications, where only 7 end-point keypoints are provided. We observe that our model outperforms other solutions while maintaining a reasonable accuracy.

\subsubsection{Qualitative Assessment} 
We present a qualitative assessment of our model to Out-of-Distribution (OoD) data to highlight our model's robustness and generalization. \Cref{fig:fig7_rich} demonstrates the visual quality of SkelFormer on RICH \cite{huang2022capturing} dataset in equally-sampled frames from the testing set. Our method yields the correct pose and shape with high overlaps with the subject. Some failure cases, in the first sequence when two of four cameras are occluded, are also demonstrated where the network has failed to predict the correct shape or pose.

In \cref{fig:fig6_qualitative}, we showcase the visual fidelity of SkelFormer by presenting more images on public datasets and in-house recording sessions using 6 GoPro cameras. Our collected videos include a running sequence in a large volume and a dying and sitting sequence in a small volume. 


\Cref{fig:fig4_amass} visualizes SkelFormer's fitting capabilities in noisy and occlusion experiments on the sequences with its highest error from AMASS \cite{mahmood2019amass} testing set. Interestingly, in very noisy circumstances, the model tries to predict a plausible pose while adhering to the input as much as possible. It also generates viable and relaxed poses in the end-point experiments. Consistent with our results, we see almost no changes to the prediction in the presence of occlusions. Please refer to our video demo to see more animations containing noise and occlusion experiments and a comparison to VPoser-t, which is a temporal optimization method.

\section{Limitations and Future Work}
The major limitation of this work is the accumulation of errors in our backbones. Although our network mitigates jitters and occlusions better than other approaches, it is still prone to artifacts caused by mediolateral mix-ups and incorrect tracking. It also produces jitters in the presence of sudden high-level occlusions. One solution to remedy such issues is to use more accurate human trackers \cite{bridgeman2019multi,ren2015faster}, 2D keypoint estimators, and triangulation techniques. However, while this can be potentially effective in most cases, it does not address the main problem at a fundamental level. Another approach could be to use temporal information to understand anomalies and noisy keypoints via more sophisticated networks. To this end, SkelFormer can be used to initialize generative models such as DMMR \cite{huang2021dynamic} and HuMoR \cite{rempe2021humor} to refine the motions.

\section{Conclusion}
In this paper, we presented SkelFormer, a novel multi-stage pipeline consisting of keypoint estimators and a skeletal transformer for markerless human motion capture. Our method leverages large amounts of motion capture data to address the poor generalization of multi-view human shape and pose estimation approaches while outperforming optimization approaches in accuracy. Through extensive experiments, we demonstrated the effectiveness of SkelFormer in several challenging conditions, including InD and OoD settings. Specifically, We achieve the best results in our InD experiments among prior multi-view pose and shape estimation approaches and show competitive performance in OoD settings, demonstrating the generalization of our pipeline. Furthermore, we show the effectiveness of the proposed elements in our skeletal transformer through our ablation study. Finally, we show that our single-frame skeletal transformer exhibits higher robustness to noise and occlusion compared to optimization approaches that rely on temporal data.

\section*{Acknowledgment}
This work was partially funded by Mitacs through the Accelerate program.

\bibliographystyle{IEEEtran}
\bibliography{main_arxiv.bib}

\end{document}